\documentclass[sigconf, screen]{acmart}
\AtBeginDocument{%
  }

\definecolor{cvprblue}{rgb}{0.21,0.49,0.74}
\usepackage{booktabs}  
\usepackage{tabularx}  
\usepackage{booktabs}
\usepackage{tabularx}
\usepackage{multirow}
\usepackage[table]{xcolor}
\usepackage{siunitx}   

\begin{document}

\title{InCarEmo: A Multimodal Dataset for In-Cabin Emotion Recognition and Driver State Monitoring}

\renewcommand{\shortauthors}{Yang et al.}

\author{Hao Yang}
\affiliation{%
  \institution{Harbin Institute of Technology}
  \city{Harbin}
  \country{China}}
\email{hyang@ir.hit.edu.cn}

\author{Yanyan Zhao}
\authornote{Corresponding author.}
\affiliation{%
  \institution{Harbin Institute of Technology}
  \city{Harbin}
  \country{China}}
\email{yyzhao@ir.hit.edu.cn}

\author{Kewei Zhao}
\affiliation{%
  \institution{Harbin Institute of Technology}
  \city{Harbin}
  \country{China}}
\email{kwzhao@ir.hit.edu.cn}

\author{Hongbo Zhang}
\affiliation{%
  \institution{Harbin Institute of Technology}
  \city{Harbin}
  \country{China}}
\email{hbzhang@ir.hit.edu.cn}

\author{Tian Zheng}
\affiliation{%
  \institution{Harbin Institute of Technology}
  \city{Harbin}
  \country{China}}
\email{tzheng@ir.hit.edu.cn}

\author{Yusheng Liu}
\affiliation{%
  \institution{Harbin Institute of Technology}
  \city{Harbin}
  \country{China}}
\email{ysliu@ir.hit.edu.cn}

\author{Xing Fu}
\affiliation{%
  \institution{Harbin Institute of Technology}
  \city{Harbin}
  \country{China}}
\email{xfu@ir.hit.edu.cn}

\author{Bichen Wang}
\affiliation{%
  \institution{Harbin Institute of Technology}
  \city{Harbin}
  \country{China}}
\email{bcwang@ir.hit.edu.cn}

\author{Yu Zhang}
\affiliation{%
  \institution{Harbin Institute of Technology}
  \city{Harbin}
  \country{China}}
\email{yzhang@ir.hit.edu.cn}

\author{Hao He}
\affiliation{%
  \institution{SERES}
  \city{Chongqing}
  \country{China}}
\email{hv1121@163.com}

\author{Zhen Wu}
\affiliation{%
  \institution{SERES}
  \city{Chongqing}
  \country{China}}
\email{15923218121@163.com}

\author{Xuda Zhi}
\affiliation{%
  \institution{SERES}
  \city{Chongqing}
  \country{China}}
\email{zhixuda_0@163.com}

\author{Yongbo Huang}
\affiliation{%
  \institution{SERES}
  \city{Chongqing}
  \country{China}}
\email{764599693@qq.com}

\author{Bing Qin}
\affiliation{%
  \institution{Harbin Institute of Technology}
  \city{Harbin}
  \country{China}}
\email{qinb@ir.hit.edu.cn}

\begin{abstract}
 Understanding driver emotion and state is critical for the next generation of intelligent in-cabin systems that ensure safety and enhance human–vehicle interaction. However, existing public datasets for in-cabin affective computing are largely limited to visual modalities and rarely include conversational information, making it difficult to capture the linguistic and interactive cues underlying driver emotion. To address these gaps, we introduce InCarEmo, a multimodal dataset for in-cabin emotion recognition and driver state monitoring. InCarEmo integrates RGB and infrared video, in-cabin audio, and dialogue text collected from scripted in-cabin scenarios designed to simulate realistic driver behaviors, covering diverse lighting conditions and driving contexts. The dataset supports three primary tasks: 1) multimodal emotion recognition, 2) fatigue detection, and 3) distraction monitoring. 
 In addition to the original Chinese data, we construct an auxiliary English benchmark to support preliminary cross-lingual evaluation.
 We provide a unified benchmark with extensive baseline results across unimodal and multimodal methods, including analyses under modality-missing and noise conditions. Experimental results demonstrate the benefits of multimodal fusion and reveal remaining challenges under real-world noise and low-light conditions. By releasing InCarEmo, we aim to establish a comprehensive foundation for robust, interpretable, and human-centric in-cabin affective understanding, promoting safer and more empathetic driver–vehicle interaction.
\end{abstract}






\renewcommand{\shortauthors}{Yang et al.}
\maketitle

\section{Introduction}

With the rapid development of intelligent vehicles and human--machine interaction systems, understanding driver emotion and state has become increasingly important for both driving safety and user experience. Accurate perception of driver emotions, fatigue, and attention can support advanced driver-assistance systems (ADAS) in reducing safety risks, mitigating distraction, and enabling more adaptive in-cabin interaction~\cite{brookhuis2001behavioural, zepf2020driver}.

Despite recent progress in affective computing and multimodal learning~\cite{yang2025large, zadeh2017tensor, tsai2019multimodal}, research on in-cabin emotion recognition and driver state monitoring is still constrained by the lack of suitable public datasets. Existing in-cabin datasets are often limited in scale, modality diversity, or environmental complexity. Many focus primarily on facial videos or physiological signals~\cite{park2016driver, massoz2016ulg}, while others are designed mainly for fatigue detection~\cite{angkititrakul2007utdrive, binas2017ddd17}. 
More importantly, existing public datasets for in-cabin affective computing are largely limited to visual modalities and rarely include conversational information, making it difficult to capture the linguistic and interactive cues underlying driver emotion. Even when dialogue is available in broader multimodal emotion datasets such as IEMOCAP~\cite{busso2008iemocap} and MELD~\cite{poria2018meld}, it is typically not grounded in driving-related scenarios or task-specific in-cabin topics, such as traffic conditions, navigation, vehicle status, trip coordination, or safety-critical events. As a result, these datasets fail to reflect the domain-specific conversational contexts in which driver emotions naturally arise and evolve.

To address these limitations, we introduce InCarEmo, a multimodal dialogue dataset for in-cabin scenarios. InCarEmo integrates RGB and infrared (IR) videos, in-cabin audio, and dialogue text collected from scripted yet realistic in-cabin scenarios. The dataset is designed to capture diverse driver-related affective and behavioral cues under varying illumination, viewpoints, and noise conditions. It supports three closely related tasks: multimodal emotion recognition, fatigue detection, and distraction monitoring. To facilitate cross-lingual research, we further provide bilingual Chinese--English textual annotations and an English benchmark setting derived from the original data. 

In addition to releasing the dataset, we conduct extensive experiments with representative unimodal and multimodal baselines, and further introduce CAMEL, a lightweight multimodal baseline tailored for in-cabin emotion recognition and driver state monitoring. Our evaluations cover challenging settings such as missing modalities, enabling a comprehensive analysis of model robustness in in-cabin scenarios. Experimental results show that multimodal fusion consistently improves performance, and that CAMEL achieves strong performance, especially under low-light and noisy conditions where single modalities are often less reliable.

In summary, our main contributions are as follows:
1) We present InCarEmo, a Chinese-centered multimodal dataset for in-cabin emotion recognition and driver state monitoring, covering RGB and IR videos, audio, and text in in-cabin scenarios, and additionally derive an auxiliary English benchmark for preliminary cross-lingual evaluation.
2) We conduct extensive experiments with representative unimodal and multimodal baselines, including evaluations under missing-modality settings.
3) We provide CAMEL, a lightweight multimodal baseline system for resource-constrained in-cabin applications, demonstrating strong recognition performance.

\section{Related Work}
\label{sec:related}
\subsection{In-cabin Emotion Recognition Datasets}
Recent in-cabin datasets such as DMD~\cite{Ortega_2020} and DriverMVT~\cite{othman2022drivermvt} primarily focus on driver-state monitoring, capturing multimodal signals (e.g., RGB, physiological) under realistic driving scenarios. Although these datasets contribute to understanding driver behavior, they lack explicit emotion annotations. More recent datasets, such as MLI-DER~\cite{yang2023robust}, On-Road DFE~\cite{xiao2022road}, and KMU-FED~\cite{jeong2018driver}, have introduced emotion recognition tasks and provided emotion labels for drivers, yet they still do not extend to in-cabin conversational contexts. With the continuous advancement of intelligent driving systems, understanding the driver’s psychological and emotional states has become increasingly crucial for ensuring safety, comfort, and human–vehicle trust. To bridge this gap, we propose InCarEmo, a in-cabin emotional dialogue dataset featuring natural driver–assistant interactions with multimodal signals, fine-grained emotion labels, and diverse real-world driving scenarios.

\subsection{Fatigue and Distraction Detection Dataset}
In recent years, driver fatigue detection has become one of the key tasks in intelligent driving. In-cabin datasets such as YawDD~\cite{10.1145/2557642.2563678} and DMD~\cite{Ortega_2020} record drivers' facial and behavioral cues (e.g., eye closure, head pose, yawning) under real or simulated driving conditions. Similarly, publicly available datasets such as 3MDAD~\cite{JEGHAM2020115960} further incorporate multimodal signals, including infrared and depth information, to enhance the robustness of fatigue and distraction recognition models. 
However, practical applications, especially in diverse or non-Western settings, remain limited, as most existing datasets primarily feature Western participants, restricting model generalization across cultures and real-world intelligent driving scenarios. 
To address these limitations, we construct an extended fatigue detection dataset based on InCarEmo to improve the generalizability and inclusiveness of driver monitoring research.

\subsection{Multimodal Emotion Recognition Dataset}
The field of multimodal emotion recognition has seen the emergence of several representative datasets. These include IEMOCAP~\cite{busso2008iemocap}, which is based on dyadic interactions between actors; CMU-MOSI~\cite{zadeh2016mosi} and CMU-MOSEI~\cite{zadeh2018multimodal}, which focus on the intensity of sentiment in video reviews; MELD~\cite{poria2018meld}, which is oriented towards multi-party conversations; and the MER2025~\cite{lian2025mer2025affectivecomputing}, which emphasizes the challenges of real-world scenarios. These works primarily cater to general-purpose or "in-the-wild" environments, concentrating on modeling emotion or sentiment polarity through the fusion of text, audio, and visual modalities. However, they do not address the specific challenges inherent to the intelligent cockpit environment. 
In contrast, our proposed \textbf{InCarEmo} dataset is specifically tailored to the intelligent cockpit setting, addressing the unique challenges of in-vehicle interactions and enabling the study of multimodal emotion recognition and driver state monitoring in in-cabin environments.

\section{InCarEmo Dataset}

\begin{figure*}[t]
\centering
\includegraphics[width=0.9\textwidth]{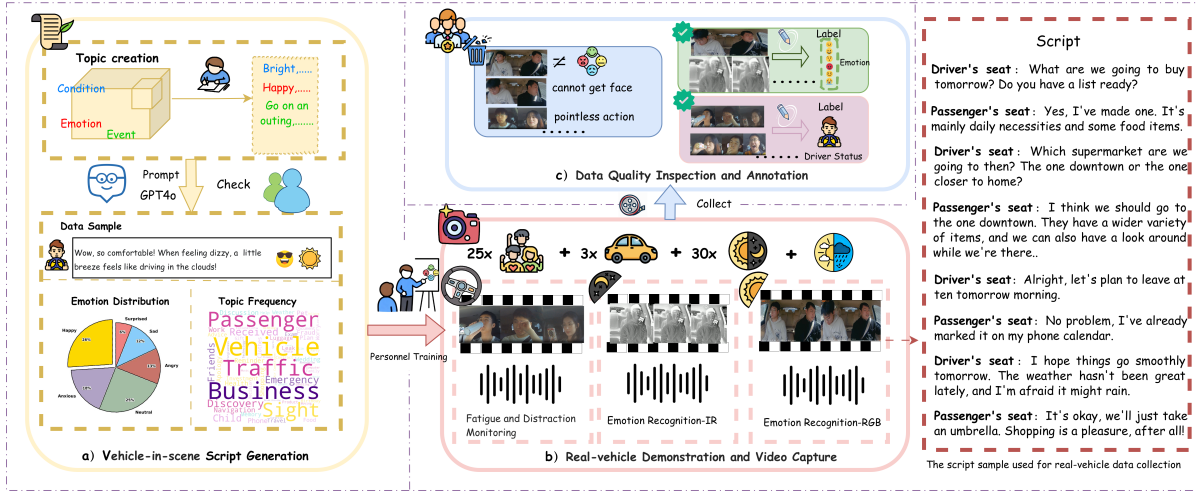}
	\caption{Overview of the InCarEmo dataset construction process.}
	\label{fig:dataset}
\end{figure*}

We introduce InCarEmo, a in-cabin multimodal emotion and safety monitoring dataset captured in realistic in-cabin environments. To the best of our knowledge, InCarEmo is the first dataset that simultaneously incorporates RGB, infrared (IR), and textual dialogue modalities for comprehensive driver emotion understanding in the in-cabin conversational setting. Figure~\ref{fig:dataset} illustrates the InCarEmo dataset construction process. 

\subsection{Collection Setting}
We captured videos of drivers expressing emotions while speaking using the front-facing camera of a VantrueCam dashcam, recording at a resolution of 1920×1080 to ensure clear facial details and stable image quality under in-cabin lighting conditions. In addition, an external microphone was used to record high-fidelity conversational audio, captured synchronously with the video to ensure multimodal data alignment. The camera was installed beneath the vehicle’s rearview mirror, ensuring clear capture of facial features and details of both the driver and front passenger.
\subsection{Collection Pipeline}
\textbf{Script Generation.}
As illustrated in part a of Figure~\ref{fig:dataset}, to ensure the authenticity and diversity of conversational coverage, we manually designed topics related to traffic conditions, trip planning, daily conversations, and vehicle status discussions, common themes in in-car dialogue. 
Based on these manually defined topics, we used GPT-4o~\cite{openai2024gpt4o} to generate approximately 2,000 multi-turn dialogue scripts that reflect natural interactions between drivers and passengers or between drivers and in-vehicle assistants. 
Each dialogue was annotated with fine-grained emotional labels by a LLM to capture subtle emotional states. 
To ensure the rationality of the dialogue scripts and the consistency of the corresponding emotions, all scripts and emotion labels were reviewed by well-trained annotators.

\textbf{Video Collection.}
We collected the data in realistic in-cabin environments with vehicles parked to ensure safety while preserving authentic lighting and cabin conditions. Each scenario involved two participants, a driver and a passenger (assistant), who spoke according to predefined scripts (we show an example in the ``Script column'' of Figure~\ref{fig:dataset}). Prior to recording, all participants underwent performance training to ensure the production of emotion-annotated dialogue videos that meet our quality standards. 
As shown in part b of Figure~\ref{fig:dataset}, we constructed 3.6K RGB and IR video clips, each capturing a complete utterance with rich facial expressions, gestures, and vocal cues. Meanwhile, some individuals participated in additional performance training to record 500 video clips of drivers, each comprehensively showcasing the driver's state and behavior.

\textbf{Expert Annotation.}
The final step, shown in part c of Figure~\ref{fig:dataset}, is quality inspection and annotation. Expert annotators evaluated each video for speech clarity, facial visibility, and consistency between the expressed emotions and the intended labels. Clips that did not meet the quality standards, such as those with unclear articulation, occluded faces, or inconsistent emotional expressions, were re-recorded to ensure data reliability and annotation accuracy. Each sample was independently annotated by three experts, and the inter-annotator agreement achieved a Cohen’s kappa coefficient of 0.87. We assign the final emotion or state label to each clip that passes the quality inspection.

\subsection{Dataset Statistics}

\begin{figure}[t]
    \centering
    \includegraphics[width=0.8\columnwidth]{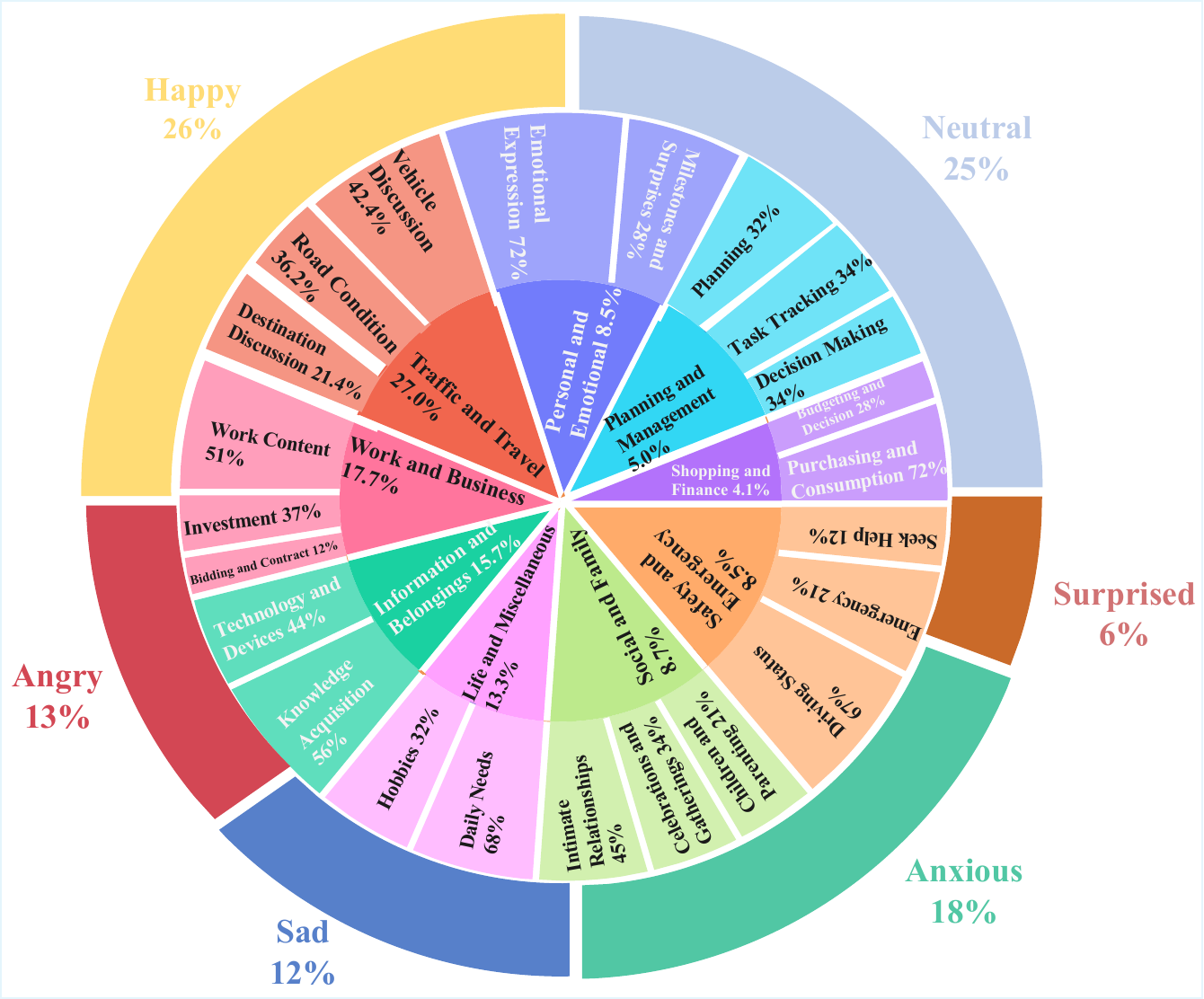}
    \caption{Emotion and topic distributions of the dataset.}
    \label{fig:dataset statistic}
\end{figure}

Figure~\ref{fig:dataset statistic} summarizes the distribution of emotion categories and scenario types. The dataset covers six target emotion classes as well as a diverse set of in-cabin conversational scenarios. A total of 25 participants were recruited to ensure diversity in facial characteristics, speaking styles, and behavioral patterns.
Table~\ref{tab:dataset statistic} reports the statistics for the three tasks.
For the emotion recognition task, we collected 3.6K emotional video clips with an average duration of 7.5 seconds, resulting in a total of 7.5 hours of multimodal recordings. To further enhance the applicability of the dataset, we additionally constructed an English version based on the original data.
In addition, the dataset contains fatigue and distraction samples to support driver state monitoring tasks.
\begin{table}
\centering
\caption{InCarEmo Dataset Statistics.}
\setlength{\tabcolsep}{5pt}  
\scalebox{0.7}{
\begin{tabular}{@{}l c c c c c@{}}  
\hline \hline
\textbf{Task} & \textbf{Lang.} & \textbf{Num.} & \textbf{Dur.(h)} & \textbf{Clip Dur.(s)} & \textbf{Percent} \\ 
\hline \hline
Emotion Recognition & Chinese & 3,600 & 7.50 & 7.5 & 58.8\% \\
Emotion Recognition & English & 2,000 & 4.20 & 7.6 & 32.7\% \\
Fatigue Detection & -- & 230 & 0.44 & 6.9 & 3.8\% \\
Distraction Detection & -- & 270 & 0.53 & 7.0 & 4.4\% \\ \hline
Total &  & 6,100 & 12.67 & 7.5 & 100.0\% \\
\hline \hline 
\end{tabular} 
}
\label{tab:dataset statistic}
\end{table}

\section{Tasks and Benchmark Definition}
The InCarEmo dataset supports three core tasks: multimodal emotion recognition, fatigue detection, and distraction monitoring.
All tasks are defined on synchronized multimodal segments. Let $V_t$, $A_t$, and $T_t$ denote the video, audio, and text within the time window $[t-w, t]$, respectively. The multimodal input is defined as
$
X_t = (V_t, A_t, T_t),
$
and the goal is to learn a task-specific mapping
$
f_\theta: X_t \rightarrow y_t,
$
where $y_t$ denotes the corresponding label.

\subsection{Multimodal Emotion Recognition}
The multimodal emotion recognition task aims to classify the driver's emotional state from synchronized visual, audio, and textual inputs under diverse in-cabin conditions. Specifically, the visual input may include RGB and IR streams. The task predicts one of six emotion categories:
$
\mathcal{C} = \{\text{neutral},\, \text{happy},\, \text{sad},\, 
\text{angry}, \text{surprised},\, \\ \text{anxious}\}.
$
Formally, the objective is to learn a classifier
$
f_{\theta}: X_t \rightarrow y^{\text{emo}}_t,\quad y^{\text{emo}}_t \in \mathcal{C},
$
where $y^{\text{emo}}_t$ is the ground-truth emotion label.

\subsection{Fatigue and Distraction Monitoring}
The fatigue and distraction monitoring task aims to recognize the driver's attention state from multimodal inputs. It covers fatigue-related behaviors such as eye closure and yawning, as well as distraction-related behaviors such as drinking and mobile phone use. We consider either a coarse-grained label space
$
\mathcal{S} = \{\text{fatigue}, \\ \text{distraction}, \, \text{normal}\},
$
and a fine-grained label space
$
\mathcal{S} = \{s_1, \dots, s_K\},
$
where each $s_k$ denotes a specific driver behavior, such as eye closure, yawning, drinking, phone usage, or smoking. The objective is to learn a classifier
$
f_{\theta}: X_t \rightarrow y^{\text{state}}_t,\quad y^{\text{state}}_t \in \mathcal{S}.
$

\subsection{Auxiliary English Benchmark}
To enhance the dataset's international applicability and facilitate broader research collaboration, we constructed an English version of our dataset. The construction process involved two main steps. First, all Chinese textual annotations were translated into English using LLM API. Subsequently, we employed the higgs-audio~\cite{higgsaudio2025} framework to synthesize corresponding English speech, using the original Chinese audio and the translated English text as inputs. Recognizing that cross-lingual speech synthesis can introduce artifacts and inaccuracies, we implemented a rigorous filtering protocol to ensure the quality of the synthesized audio:
\textbf{Validity Check.} 
We began by removing samples where the synthesis process failed outright or produced invalid outputs.
\textbf{Content Consistency Verification.} 
We utilized an Automatic Speech Recognition model to transcribe the synthesized English audio. Samples were discarded if the Word Error Rate between the ASR output and the target English text exceeded a predefined threshold.
\textbf{Duration Alignment Analysis.} 
We compared the duration of the synthesized English audio with the original Chinese audio, filtering out instances with significant length discrepancies, which often indicate synthesis failures.
This multi-stage filtering process resulted in the exclusion of approximately 40\% of the synthesized audio samples, retaining only those of higher quality for the benchmark evaluation.

\section{Method}
Beyond introducing the InCarEmo dataset, we further propose \textbf{C}ontrastive \textbf{A}lignment and \textbf{M}ulti-classifier \textbf{E}nsemble \textbf{L}earning as baseline frameworks for the dataset, including \textbf{CAMEL-emotion} for multimodal emotion recognition and \textbf{CAMEL-state} for driver state monitoring.
These baselines are specifically designed to operate efficiently in resource-constrained in-vehicle environments while maintaining competitive recognition performance. In this section, we provide a detailed description of the model architectures and training strategies used for each task.

\subsection{Multimodal Emotion Recognition}

The CAMEL-emotion is a lightweight, real-time model designed for robust emotion recognition inside vehicular cabins. 
Given synchronized video, audio, and textual dialogue streams within the time window $[t-w, t]$, we denote the corresponding visual clip, audio waveform, and transcript as $V_t$, $A_t$, and $T_t$, and define the multimodal input as
$
X_t = (V_t, A_t, T_t).
$
In all experiments, the temporal window is fixed to $w=5$ seconds to balance temporal coverage and computational efficiency.
Each modality is processed by an efficient pretrained backbone to obtain compact unimodal representations:
\begin{equation}
z_t^{v} = f^{v}(V_t), \quad 
z_t^{a} = f^{a}(A_t), \quad 
z_t^{t} = f^{t}(T_t), 
\end{equation}
where $f^{v}$ is based on CLIP for RGB/IR and thermal inputs, $f^{a}$ is a Chinese-HuBERT encoder, and $f^{t}$ is a Qwen3-based text encoder.

\textbf{Stage 1: Modality-Specific Fine-Tuning.}
In the first stage, we fine-tune each encoder separately on the in-cabin emotion recognition task so that it learns emotion-sensitive features for its modality.

We adopt Qwen3-0.6B ~\cite{yang2025qwen3} as the text encoder and perform supervised fine-tuning on a small emotion-centric dialogue set to enhance sensitivity to affective linguistic cues. 
For the audio modality, we fine-tune Chinese-HuBERT ~\cite{TencentGameMate} on 16\,kHz speech segments of 5 seconds. 
A lightweight classification head is appended to capture rhythm, energy, and pitch variations correlated with emotional states.
For RGB inputs, we employ \texttt{face\_recognition}~\cite{king2009dlib,geitgey2017facerecognition} to detect driver and passenger faces every five frames, concatenate the cropped faces along the temporal dimension, and feed them into a CLIP-base ViT backbone ~\cite{radford2021learning}. 
To exploit complementary temperature cues, we detect thermal face crops at the same sampling rate, and feed them into a CLIP-Large backbone. 
Both visual streams are fine-tuned with an emotion classification loss, enabling the model to capture subtle facial dynamics and heat-distribution patterns related to arousal and stress.

\textbf{Stage 2: Cross-Modal Contrastive Alignment.}
Inspired by the modality representation contrastive learning strategy of Fan et al.~\cite{fan2024leveraging}, we introduce a cross-modal contrastive module in the second stage to further align and fuse multimodal emotion representations. 
We aim to learn a shared emotional subspace in which embeddings from different modalities of the same segment are pulled closer, while embeddings from different segments are pushed apart.

Concretely, for modality pairs $(m,n) \in \{(v,a), (v,t), (a,t)\}$, we define a cross-modal contrastive loss:
\begin{equation}
\mathcal{L}_{mn}
=
-\sum_{t}
\log
\frac{
\exp\big(\mathrm{sim}(z_t^{m}, z_t^{n}) / \tau\big)
}{
\sum\limits_{j}
\exp\big(\mathrm{sim}(z_t^{m}, z_j^{n}) / \tau\big)
},
\end{equation}
where $\mathrm{sim}(\cdot,\cdot)$ denotes cosine similarity and $\tau$ is a temperature hyperparameter. 
The overall cross-modal alignment objective is
\begin{equation}
\mathcal{L}_{\text{cross}}
=
\mathcal{L}_{va}
+
\mathcal{L}_{vt}
+
\mathcal{L}_{at},
\end{equation}
which jointly constrains the projection spaces of text, audio, and visual modalities.

\textbf{Stage 3: Prediction with Multi-Classifier Voting.}
On top of the aligned embeddings, several lightweight classifiers are trained for different modality combinations (e.g., $v$, $a$, $t$, $va$, $vat$). 
At inference time, we aggregate their output probabilities via soft voting to obtain the final emotion prediction. 
This multi-classifier voting scheme improves robustness under noisy in-cabin conditions while maintaining low latency suitable for real-time deployment.


\subsection{Driver State Monitoring}

\paragraph{Fatigue Detection.}
For fatigue analysis, we employ the lightweight MediaPipe FaceMesh model ~\cite{lugaresi1906mediapipe} to extract dense 3D facial landmarks around the eyes and mouth. 
Using these landmarks, we compute the eye aspect ratio (EAR), mouth aspect ratio (MAR), and the percentage of eye closure (PERCLOS), which serve as physiological indicators of drowsiness:
\begin{equation}
\mathrm{EAR} = 
\frac{\|p_2 - p_6\| + \|p_3 - p_5\|}
{2 \, \|p_1 - p_4\|},
\mathrm{MAR} =
\frac{\|q_3 - q_9\|}{\|q_1 - q_5\|},
\end{equation}
where $p_i$ and $q_j$ denote the eye and mouth landmark coordinates. 
PERCLOS is computed as the proportion of frames in which EAR falls below a predefined threshold. 
high PERCLOS indicates prolonged eye closure, while high MAR identifies yawning behavior. 
These features are fed into a lightweight classifier to determine whether the driver is exhibiting fatigue.

\paragraph{Distraction Recognition.}
For distraction monitoring, we employ CLIP's visual encoder to extract semantic visual representations of driver actions.
Given a visual clip $V_t$, the encoder produces a feature embedding:
\begin{equation}
z^{v}_t = f^{v}(V_t),
\end{equation}
which is then passed through a task-specific classification head:
\begin{equation}
\hat{y}^{\text{dis}}_t = g_{\phi}(z^{v}_t).
\end{equation}
This classifier is trained using cross-entropy supervision and is designed to identify common in-vehicle distracted behaviors.

\section{Experiment}
\subsection{Experimental Setup}
\begin{figure}[t]
    \centering
    \includegraphics[width=\columnwidth]{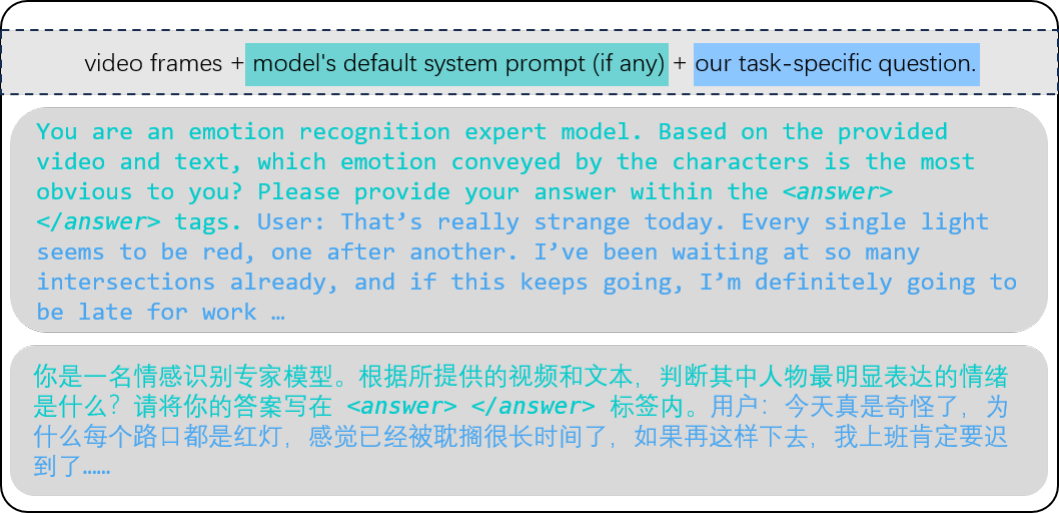}
    \caption{An example of prompting a Video LLM.}
    \label{fig:prompt}
\end{figure}

We adopt a unified evaluation protocol for all tasks. The dataset is split in a subject-independent manner into training, validation, and test sets with a ratio of 80\%, 10\%, and 10\%, ensuring that subjects in the test set do not appear in training. We report Accuracy, Precision, Recall, and F1-score as the main evaluation metrics. To ensure fair comparison, all models are trained and evaluated under consistent input resolutions, frame rates, and preprocessing settings.

The training procedure consists of two phases. In the first stage, we adopted CLIP-base as the vision encoder (150M), Chinese-HuBERT as the audio encoder (94M), and Qwen3-0.6B as the text encoder (596M). Each encoder is trained for 3 epochs using a learning rate of \(\mathbf{1 \times 10^{-5}}\). The training objective is the classification loss. Each encoder is trained on a single NVIDIA A100 GPU with 80GB memory. Finally, by integrating these fine-tuned encoders with the final fusion-voting network (104M), we obtained \textbf{CAMEL-emotion}(944M). In the second stage, the full model is trained for 100 epochs on a single NVIDIA A100 GPU with 40GB memory. The learning rate is set to \(\mathbf{5 \times 10^{-5}}\).
For model evaluation, we provide the model with video frames and a task-specific question, following the format: video frames + model’s default system prompt (if any) + our task-specific question as shown in Figure~\ref{fig:prompt}.

\subsection{Emotion Recognition Results on Different Modalities}

\textbf{Text Modality.} We first evaluate text-only models on InCarEmo in a zero-shot setting using transcribed in-cabin conversations as input. As shown in Table~\ref{tab:all_modality_baseline}, Qwen3-0.6B performs poorly (25.92\% Acc., 7.69\% F1), while Qwen3-8B achieves substantially better results (54.93\% Acc., 55.94\% F1). These results indicate that text alone provides limited emotional evidence for in-cabin interactions.

\textbf{Audio Modality.} We evaluate several pretrained speech models, including Chinese-HuBERT, Chinese-HuBERT-Large, and Whisper~\cite{cao2012whisper}. All encoders are frozen and equipped with a lightweight classification head. SpecAugment and random gain perturbation are applied during training. As shown in Table~\ref{tab:all_modality_baseline}, Whisper achieves the best audio-only performance (24.51\% Acc.), while Chinese-HuBERT-Large slightly outperforms Chinese-HuBERT.

\textbf{Video Modality.} We evaluate CLIP, emoCLIP~\cite{foteinopoulou2024emoclip}, and GPT-4o. For CLIP-based models, we freeze the visual encoder and train an MLP classifier; GPT-4o is evaluated in a zero-shot manner. As shown in Table~\ref{tab:all_modality_baseline}, emoCLIP outperforms standard CLIP variants, and GPT-4o achieves the best visual performance.

\textbf{Infrared Thermal Modality.} We further evaluate CLIP and LanguageBind Thermal~\cite{zhu2023languagebind} on thermal images. As shown in Table~\ref{tab:all_modality_baseline}, LanguageBind Thermal achieves strong performance (56.13\% Acc., 56.82\% F1), substantially outperforming CLIP.

\textbf{Multimodal Fusion.} We finally benchmark multimodal large models and fusion-based architectures on InCarEmo. As shown in Table~\ref{tab:all_modality_baseline}, multimodal models consistently outperform single-modality baselines, highlighting the complementary value of cross-modal cues. Among general-purpose models, GPT-4o achieves the best zero-shot accuracy (75.58\%). Open-source models such as Qwen2.5-VL~\cite{bai2025qwen2} and R1-Omni~\cite{zhao2025r1} obtain moderate results, while emotion-oriented models, including Emotion-LLaMA~\cite{cheng2024emotion} and AffectGPT~\cite{lian2024affectgpt}, bring only limited gains. Our model achieves the best overall performance with 82.25\% accuracy, demonstrating the effectiveness of our design.

\begin{table}
\centering
\footnotesize 
\setlength{\tabcolsep}{3pt} 
\caption{Performance comparison of different models on emotion recognition. Best results are bold.}
\scalebox{0.9}{
\begin{tabular}{l|c c c c c} 
\hline \hline
\textbf{Method} & \textbf{Params} & \textbf{Acc} & \textbf{Pre} & \textbf{Rec} & \textbf{F1} \\ 
\hline 
\multicolumn{6}{c}{\textit{Text-only (T)}} \\ 
\hline 
Qwen3-0.6B & 752M & 25.92 & 12.75 & 16.86 & 14.52 \\ 
Qwen3-8B & 8.19B & 54.93 & 71.73 & 54.28 & 55.94 \\ 
\hline 
\multicolumn{6}{c}{\textit{Audio-only (A)}} \\ 
\hline 
Whisper & 74M & 24.51 & 12.35 & 22.56 & 13.27 \\
Chinese-HuBERT & 95M & 16.34 & 10.52 & 15.76 & 11.22 \\ 
Chinese-HuBERT-Large & 317M & 20.32 & 11.27 & 20.34 & 12.68 \\  
\hline 
\multicolumn{6}{c}{\textit{Visual-only (V)}} \\ 
\hline 
CLIP & 151M & 24.23 & 29.62 & 28.93 & 20.43 \\ 
CLIP-Large & 428M & 19.72 & 30.95 & 28.40 & 18.00 \\ 
emoCLIP & 151.3M & 29.19 & 31.37 & 29.86 & 24.79 \\ 
GPT-4o & - & 57.56 & 59.05 & 50.95 & 50.94 \\ 
\hline 
\multicolumn{6}{c}{\textit{Infrared Thermal (IR)}} \\ 
\hline 
CLIP & 151M & 47.38 & 34.70 & 41.26 & 39.45 \\ 
LanguageBind & 330M & 56.13 & 59.08 & 52.64 & 56.82 \\ 
\hline 
\multicolumn{6}{c}{\textit{Text+Visual (T+V)}} \\ 
\hline 
GPT-4o & - & 75.58 & 76.33 & 77.21 & 75.40 \\ 
Qwen2.5-VL-3B & 3.75B & 50.87 & 61.21 & 44.69 & 46.36 \\
Qwen2.5-VL-7B & 8.29B & 55.78 & 63.42 & 55.28 & 54.55 \\
\hline 
\multicolumn{6}{c}{\textit{Multimodal (T+V+A)}} \\ 
\hline 
HumanOmni-5B ~\cite{zhao2025humanomni} & 1.37B & 22.82 & 19.78 & 19.96 & 10.75 \\ 
R1-Omni & 1.37B & 18.13 & 7.81 & 7.42 & 5.90 \\ 
+ft.     & 1.37B     & 47.91 (+29.78\%) & 42.73 & 38.54 & 31.56 (+25.66\%) \\
Emotion-LLaMA & 7B & 46.48 & 50.12 & 51.46 & 44.60 \\ 
+ft.   & 7B       & 74.65 (+28.17\%) & 73.63 & 73.05 & 73.13 (+28.53\%) \\
AffectGPT & 7B & 57.75 & 57.92 & 55.41 & 53.54 \\ 
+ft.     & 7B     & 73.24 (+15.49\%) & 71.63 & 74.54 & 72.50 (+18.96\%) \\
\rowcolor{gray!10}
\textbf{CAMEL-emotion} & 944M & \textbf{82.25} & \textbf{80.95} & \textbf{78.52} & \textbf{79.56} \\ 
\hline \hline 
\end{tabular} 
}
\label{tab:all_modality_baseline}
\end{table}

\subsection{Fatigue and Distraction Monitoring Results}
We rigorously evaluated data covering various driver states, focusing on two critical behavior categories: fatigue (e.g., dozing, yawning) and distraction (e.g., drinking, smoking). We assessed several multimodal large models on this subset, including Qwen2.5-VL-3B, Qwen2.5-VL-7B, and GPT-4o, alongside a small model trained specifically on this dataset. As shown in Table~\ref{tab:detection_results}, our model achieved 84.58\% accuracy on fatigue detection. Qwen2.5-VL-7B attained the highest zero-shot accuracy (66.08\%), demonstrating strong generalization, whereas Qwen2.5-VL-3B and GPT-4o underperformed, with F1 scores below 40\%. For distraction detection, Qwen2.5-VL-7B again led with 61.17\% zero-shot accuracy, while Qwen2.5-VL-3B and GPT-4o showed limited effectiveness.

\begin{table}[t]
\centering
\footnotesize 
\caption{Performance comparison of different models on fatigue and distraction detection tasks. Best results are bold.}
\setlength{\tabcolsep}{5pt} 
\scalebox{0.82}{
\begin{tabular}{l c c c c c c c c} 
\hline \hline
\multirow{2}{*}{\textbf{Model}} & \multicolumn{4}{c}{\textbf{Fatigue Detection}} & \multicolumn{4}{c}{\textbf{Distraction Detection}} \\
\cmidrule(lr){2-5} \cmidrule(lr){6-9}
& \textbf{Acc} & \textbf{Pre} & \textbf{Rec} & \textbf{F1} & \textbf{Acc} & \textbf{Pre} & \textbf{Rec} & \textbf{F1} \\ 
\hline \hline
Qwen2.5-VL-3B & 41.41 & 46.14 & 38.81 & 27.91 & 37.00 & 55.04 & 32.38 & 25.03 \\
Qwen2.5-VL-7B & 66.08 & 75.74 & 67.70 & 65.52 & 61.17 & 42.11 & 42.01 & 42.00 \\
GPT-4o & 37.89 & 33.56 & 21.51 & 19.97 & 49.45 & 29.74 & 32.54 & 30.22 \\
\rowcolor{gray!10}
\textbf{CAMEL-state} & \textbf{84.58} & \textbf{78.31} & \textbf{84.77} & \textbf{80.43} & \textbf{81.06} & \textbf{78.09} & \textbf{85.60} & \textbf{81.24} \\

\hline \hline 
\end{tabular} 
}
\label{tab:detection_results}
\end{table}

\subsection{Ablation Study}
\textbf{Modality Ablation.} To assess the contribution of each modality, we perform a modality ablation study in Table~\ref{tab:modal}, covering unimodal (V, A, T), bimodal (VA, VT, AT), and trimodal (VAT) settings. Among the unimodal variants, audio (A) performs best, while text (T) performs worst, suggesting that audio provides more informative cues for this task. Combining modalities consistently improves performance, with all bimodal variants surpassing unimodal ones, highlighting the benefit of cross-modal interaction. The trimodal setting (VAT) achieves the best overall results (Acc: 82.25, F1: 79.56), indicating that full multimodal integration yields the most comprehensive understanding.
\begin{table}[t]
\centering
\footnotesize 
\caption{Performance comparison between the original Chinese dataset and the synthesized English dataset across all unimodal and multimodal settings.}
\setlength{\tabcolsep}{5pt} 
\scalebox{0.89}{
\begin{tabular}{l l c c c c c c c} 
\hline \hline
\textbf{Language} &\textbf{Metrics} & \textbf{V} & \textbf{A} & \textbf{T} & \textbf{VA} & \textbf{VT} & \textbf{AT} & \textbf{VAT} \\ 
\hline \hline
\multirow{4}{*}{Chinese} & Accuracy   & 70.99 & 73.80 & 41.69 & 81.13 & 71.83& 80.00 & 82.25 \\
                          & Precision  & 68.64 & 71.81 & 28.70 & 79.65 & 72.80 & 77.23 & 80.95 \\   
                          & Recall     & 67.76 & 73.67 & 31.66 & 79.10 & 67.25 & 76.58 & 78.52 \\ 
                          & F1-score   & 67.85 & 71.50 & 28.28 & 79.15 & 68.66 & 76.79 & 79.56 \\
\hline
\multirow{4}{*}{English} &Accuracy   & 72.52 & 48.09 & 51.91 & 69.47 & 74.81 & 58.78 & 77.86 \\
&Precision     & 69.71 & 58.59 & 48.25 & 64.01 & 69.54 & 59.17 & 75.27 \\   
&Recall     & 66.96 & 43.28 & 54.15 & 61.22 & 69.98 & 61.24 & 72.66 \\ 
&F1-score    & 67.46 & 43.98 & 49.08 & 60.79 & 68.78 & 57.23 & 72.97 \\ 
\hline \hline 
\end{tabular} 
}
\label{tab:modal}
\end{table}

\begin{figure}[tp!]
\centering
\includegraphics[width=0.47\textwidth]{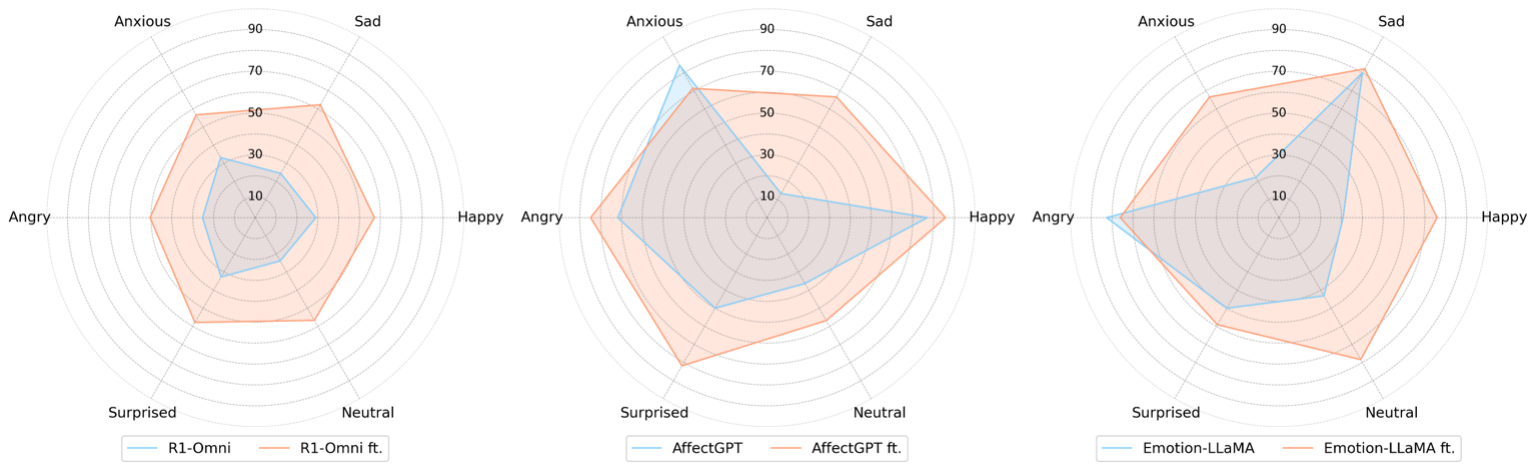}
	\caption{Comparison of each emotion category recognition accuracy fine-tuning on the InCarEmo dataset.}
	\label{fig:finetune_model}
\end{figure}

\textbf{Comparison of Fine-Tuned Multimodal Models.} 
To ensure a rigorous and fair comparison, we further fine-tune several representative multimodal emotion models on the InCarEmo dataset. 
This setting allows us to directly evaluate their performance under identical training splits, optimization configurations, and evaluation protocols, ensuring that any performance differences arise from architectural or fusion design rather than data or procedural inconsistencies. 
As presented in Table~\ref{tab:all_modality_baseline}, all models exhibit substantial improvement after fine-tuning on our dataset: Emotion-LLaMA improves to 74.65\%, AffectGPT to 73.24\%, and R1-Omni to 47.91\%. 
Figure~\ref{fig:finetune_model} further illustrates the clear performance gains achieved through task-specific adaptation.

\textbf{Auxiliary English Result Analysis.}
Results on the auxiliary English benchmark are reported only as supplemental evidence of cross-lingual transferability.
As shown in Table~\ref{tab:modal}, performance drops are mainly observed in audio-involved settings (A, VA, AT, and VAT), while the overall trends across modalities remain consistent. These results highlight the additional difficulty of cross-lingual multimodal emotion recognition, especially when emotional cues must be transferred across languages and speech conditions. At the same time, the English benchmark extends InCarEmo beyond a monolingual setting and provides a useful testbed for future research on multilingual and cross-cultural affective computing.

\section{Conclusion}
We presented InCarEmo, a comprehensive multimodal dataset for in-cabin emotion recognition and driver state monitoring, integrating RGB and infrared videos, in-cabin audio, and dialogue text from realistic driving scenarios. The dataset supports emotion, fatigue, and distraction tasks, offering a unified benchmark for multimodal in-cabin affective understanding. Extensive experiments demonstrate the benefits of multimodal fusion under diverse illumination and noise conditions, and our proposed lightweight baseline model achieves real-time performance with competitive accuracy. We expect InCarEmo to serve as a foundation for future research in affective computing and human-centered intelligent mobility, fostering safer and more empathetic in-cabin interaction.

\section*{Ethical Considerations and Data Availability}
Our data collection process strictly adheres to established ethical and privacy guidelines. All recording and annotation activities were conducted only after obtaining written informed consent from participants, who were compensated at rates exceeding local minimum-wage standards. Our quality-assurance pipeline includes multiple layers of rigorous review to identify and remove any potentially harmful, biased, or inappropriate content. These measures ensure that the resulting dataset supports fair, responsible, and trustworthy research on emotional states and driver conditions in in-cabin scenarios.
The InCarEmo dataset is publicly available at \url{https://github.com/zkw52/InCarEmo} and licensed for non-commercial academic research and use. To promote transparency and facilitate further research, we provide the open-source fine-tuning code for the CAMEL model on the InCarEmo dataset, along with the resulting fine-tuned model weights.

\bibliographystyle{ACM-Reference-Format}
\bibliography{sample-base}










\end{document}